\documentclass[conference]{IEEEtran}
\IEEEoverridecommandlockouts
\usepackage{cite}
\usepackage{amsmath,amssymb,amsfonts}
\usepackage{algorithmic}
\usepackage{graphicx}
\usepackage{textcomp}
\usepackage{xcolor}
\usepackage{multirow}
\usepackage{booktabs}
\usepackage{url}
\def\BibTeX{{\rm B\kern-.05em{\sc i\kern-.025em b}\kern-.08em
    T\kern-.1667em\lower.7ex\hbox{E}\kern-.125emX}}

\begin{document}

\title{High-Resolution Flood Extent Detection Using Deep Learning with Random Forest–Derived Training Labels\\
\thanks{\footnotesize This research was partly supported by the National Science Foundation (award number 2203180). 

Copyright 2026 IEEE. Published in the 2026 IEEE International Geoscience and Remote Sensing Symposium (IGARSS 2026), scheduled for 9 - 14 August 2026 in Washington, D.C.. Personal use of this material is permitted. However, permission to reprint/republish this material for advertising or promotional purposes or for creating new collective works for resale or redistribution to servers or lists, or to reuse any copyrighted component of this work in other works, must be obtained from the IEEE. Contact: Manager, Copyrights and Permissions / IEEE Service Center / 445 Hoes Lane / P.O. Box 1331 / Piscataway, NJ 08855-1331, USA. Telephone: + Intl. 908-562-3966.}
}
\author{\IEEEauthorblockN{1\textsuperscript{st} Azizbek Nuriddinov}
\IEEEauthorblockA{\textit{Department of Civil Engineering} \\
\textit{Florida State University}\\
Tallahassee, FL, USA}
\and
\IEEEauthorblockN{2\textsuperscript{nd} Ebrahim Ahmadisharaf}
\IEEEauthorblockA{\textit{Department of Civil Engineering} \\
\textit{Florida State University}\\
Tallahassee, FL, USA}
\and
\IEEEauthorblockN{3\textsuperscript{rd} Mohammad Reza Alizadeh}
\IEEEauthorblockA{\textit{Dept. of Biosystems \& Agricultural Engineering} \\
\textit{Michigan State University}\\
East Lansing, MI, USA}
}

\maketitle

\begin{abstract}
Validation of flood models, used to support risk mitigation strategies, remains challenging due to limited observations during extreme events. High-frequency, high-resolution optical imagery ($\sim$3 m), such as PlanetScope, offers new opportunities for flood mapping, although applications remain limited by cloud cover and the lack of labeled training data during disasters. To address this, we develop a flood mapping framework that integrates PlanetScope optical imagery with topographic features using machine learning (ML) and deep learning (DL) algorithms. A Random Forest model was applied to expert-annotated flood masks to generate training labels for DL models, U-Net. Two U-Net models with ResNet18 backbone were trained using optical imagery only (4 bands) and optical imagery combined with Height Above Nearest Drainage (HAND) and topographic slope (6 bands). Hurricane Ida (September 2021), which caused catastrophic flooding across the eastern United States, including the New York City metropolitan area, was used as an example to evaluate the framework. Results demonstrate that the U-Net model with topographic features achieved very close performance to the optical-only configuration (F1=0.92 and IoU=0.85 by both modeling scenarios), indicating that HAND and slope provide only marginal value to inundation extent detection. The proposed framework offers a scalable and label-efficient approach for mapping inundation extent that enables modeling under data-scarce flood scenarios.
\end{abstract}


\section{Introduction}

Artificial Intelligence models have transformed many areas of water research ~\cite{rabby2026mlwq,imtiaz2025simclr,ali2026ecoli,alamdari2026algalblooms}, and validated flood models are vital for infrastructure design, emergency management, and risk mitigation. Validation is challenging due to limited observations during extreme events and coarse resolution of available data. In operational scenarios, authorities and responders require rapid flood maps during ongoing hazard events when independent ground truth products are typically unavailable. Satellite observations play a critical role, yet their effectiveness is limited by trade-offs in spatial resolution, temporal frequency, and coverage. Commercial high-resolution sensors like PlanetScope (near-daily 3 m) provide spatial detail necessary for detailed flood mapping but face coverage limitations \cite{b1}. Hurricane Ida made landfall in Louisiana on August 29, 2021, as a Category 4 storm, then tracked northeast. Reaching New York City on September 1-2, it produced 180-250 mm of rainfall in 24 hours, overwhelming drainage infrastructure. Flash flooding inundated subway stations, roadways, and basement apartments, claiming 13 lives in NYC alone. This disaster exposed critical vulnerabilities and underscored the need for rapid, accurate flood extent mapping. Recent advances in deep learning show promise for automated flood detection. U-Net architectures leverage encoder-decoder structures with skip connections to preserve spatial context while capturing multi-scale features \cite{b2}. Recent implementations have achieved high accuracies, with studies reporting IoU scores exceeding 0.85 in flood detection \cite{b3}. Machine learning approaches like Random Forest provide competitive performance with added interpretability through feature importance analysis. IEEE research has demonstrated Random Forest effectively mapping flood extents using multi-sensor satellite imagery, achieving classification accuracies above 96\% \cite{b4}, \cite{b5}. Inland waterbodies exhibit inherently complex hydrology \cite{MITRA2025134075}; accordingly, integration of topographic features such as Height Above Nearest Drainage (HAND), which quantifies vertical distance to the nearest stream, has shown potential to improve flood susceptibility modeling \cite{b6}. HAND provides physically meaningful terrain context complementing spectral signatures by encoding drainage-based flood susceptibility \cite{b7}, \cite{b8}. Research gaps persist: (1) limited evaluation of progressive annotation strategies combining expert knowledge with machine learning for operational flood mapping, and (2) insufficient comparison between topographic-enhanced and optical-only configurations in high-resolution flood mapping. This study addresses these gaps through a multi-model framework where Random Forest-generated annotations are used to train U-Net models for Hurricane Ida flood mapping in the Lower Hudson basin, integrating 3-m PlanetScope imagery with HAND and slope topographic features.

%\footnotetext{Ebrahim Ahmadisharaf was partly supported by the National Science Foundation (award number 2203180).}

\section{Study Area and Data}

\subsection{Study Area}

The Lower Hudson basin (HUC-6) encompasses 10,062 km\textsuperscript{2} in southeastern New York and northern New Jersey, including New York City's five boroughs and surrounding suburban areas. The region exhibits diverse topography ranging from coastal lowlands at sea level to inland valleys and ridges reaching 200 meters elevation. Hurricane Ida's remnants arrived during the evening of September 1, 2021, producing heavy precipitation that exceeded 75 mm per hour in some locations, resulting in two distinct flooding patterns: coastal storm surge along tidally influenced reaches, and inland flash flooding from overwhelmed streams and drainage systems.

\begin{figure}[htbp]
\centerline{\includegraphics[width=0.40\textwidth, height=0.35\textwidth]{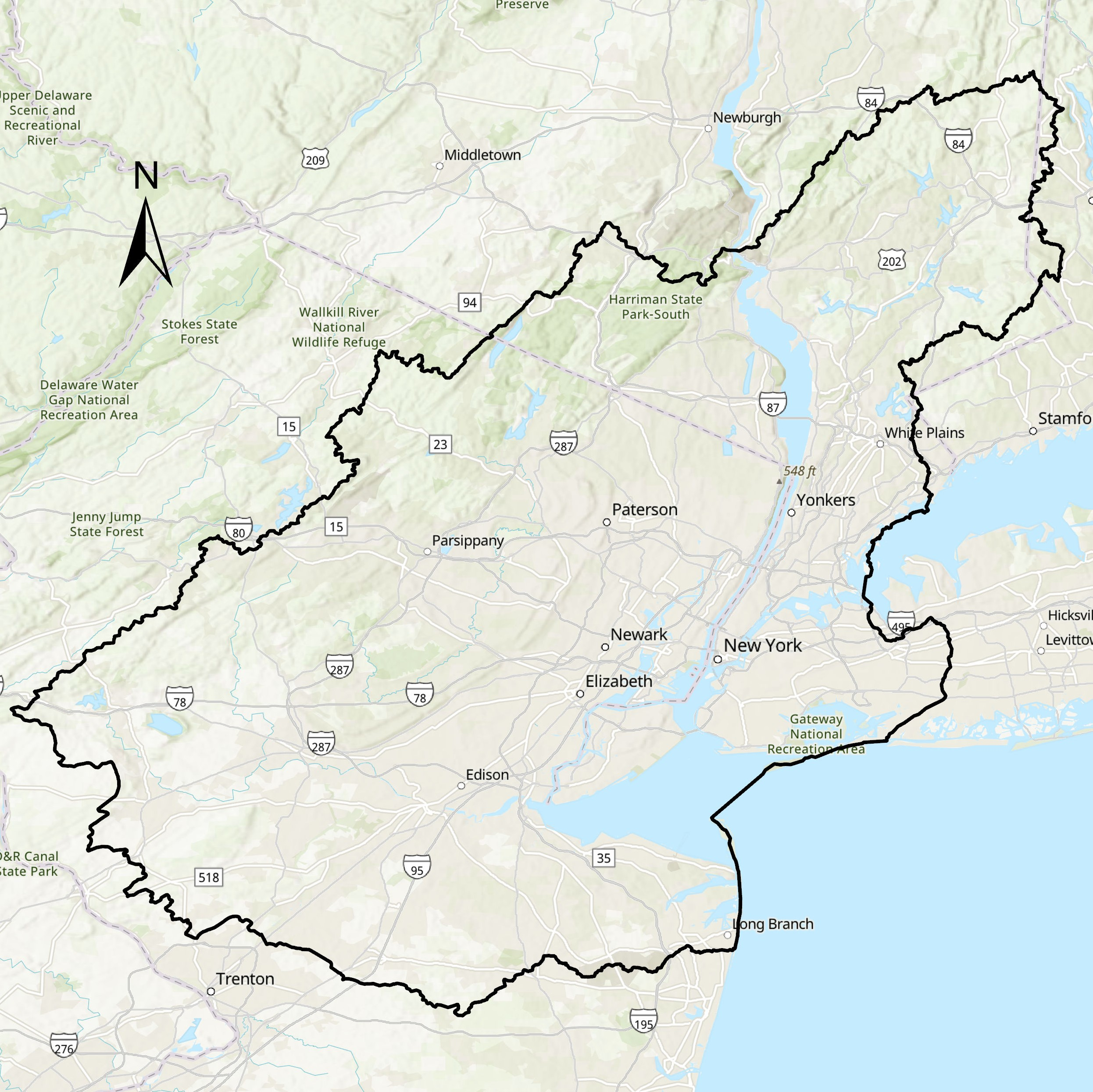}}
\caption{Study area map showing the HUC-6 Lower Hudson Basin in the northeastern United States.}
\label{fig:study_area}
\end{figure}

\subsection{Satellite and Topographic Data}

We acquired PlanetScope imagery for September 2, 2021, providing the primary high-resolution dataset for model development. Planet's 3-meter spatial resolution captures street-scale flooding patterns with four spectral bands, (Blue: 0.45-0.51 $\mu$m, Green: 0.53-0.59 $\mu$m, Red: 0.64-0.67 $\mu$m, NIR: 0.85-0.88 $\mu$m). The imagery underwent orthorectification and atmospheric correction by the data provider, with cloud cover less than 10\%. Topographic slope was derived from USGS' 10-m Digital Elevation Model. HAND quantifies vertical distance from each pixel to the nearest stream channel \cite{b6}, obtained from NHDPlus at 10-meter resolution. Both topographic layers were resampled to 3 meters using bilinear interpolation to match Planet imagery resolution.

\subsection{Ground Truth Data}

140 HWM points from USGS post-storm field surveys served as reference locations where flooding was most likely to occur and guided the spatial sampling strategy for validating data extraction.

\section{Methodology}

\subsection{Progressive Annotation Framework}

The methodology employed a three-stage progressive annotation strategy designed to overcome limitations of traditional threshold-based approaches and address the challenge of generating train data in scenarios where independent validation products are unavailable. To address the limitations of Normalized Difference Water Index (NDWI) and OTSU-derived adaptive thresholds in false detection of inundated areas, we implemented a hybrid annotation workflow. A single representative 1024$\times$1024 pixel sample containing both built-up and flooded areas was manually annotated using all available spectral bands (RGB and NIR), building coverage information, and visual cues to separate flooded from unflooded pixels. This expert-annotated sample served as training data for the RF model utilizing seven input features: four spectral bands (RGB + NIR), two topographic features (Slope + HAND), and the derived NDWI index. The trained RF achieved approximately 99\% accuracy on the annotated sample and was applied to predict flood masks for all remaining samples. These ML-predicted masks demonstrated substantial improvement over threshold-based approaches, though some isolated noisy detections persisted. These RF-generated flood masks then served as training labels for U-Net deep learning models, which were anticipated to add spatial context and reduce noise through convolutional processing.

\begin{figure}[htbp]
\centerline{\includegraphics[width=\columnwidth]{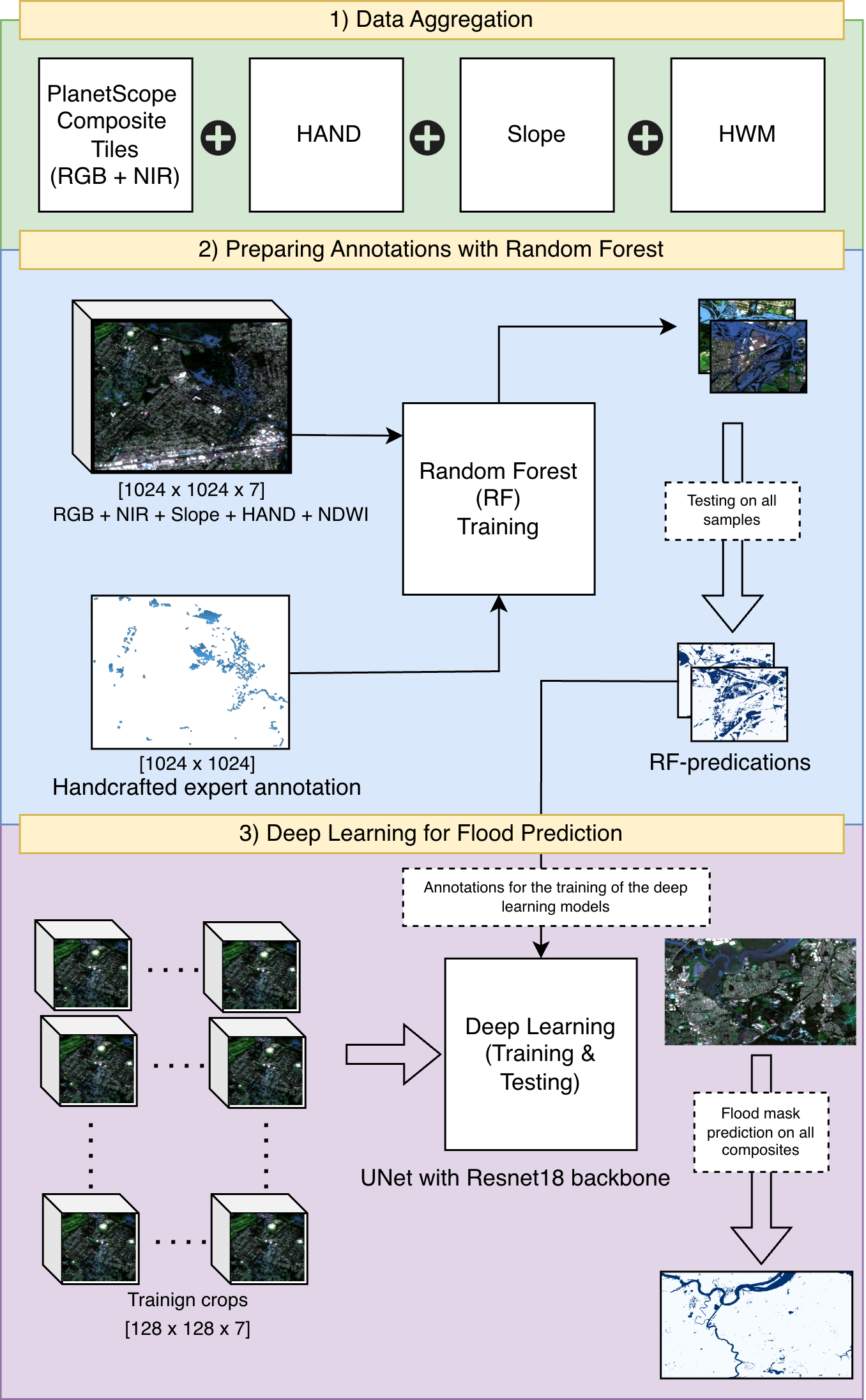}}
\caption{Methodological framework flowchart depicting the three-stage workflow. Stage 1 - Data Aggregation: PlanetScope Composite Tiles (RGB + NIR) combined with HAND, Slope, and HWM reference points. Stage 2 - Preparing Annotations with Random Forest: A 1024$\times$1024$\times$7 band cube (RGB + NIR + Slope + HAND + NDWI) containing handcrafted expert annotation serves as RF training input, producing RF predictions and flood masks tested on all samples. Stage 3 - Deep Learning for Flood Prediction: Multiple 128$\times$128$\times$7 training crops are used as annotations for training the deep learning models. UNet with ResNet18 backbone performs training and testing, generating final flood mask predictions on all composite tiles.}
\label{fig:methodology}
\end{figure}

\subsection{Spatial Sampling Strategy}

Spatial sampling was guided by HWM points from post-storm surveys. For each HWM point falling within the two PlanetScope composite scenes, a 1024$\times$1024 pixel crop was extracted. HWM points whose surrounding crops fell outside scene extents were excluded. This sampling strategy ensured training data focused on areas with confirmed flooding while capturing diverse land cover contexts.

\subsection{Feature Extraction}

Both U-Net and RF models utilized identical spectral features to enable direct comparison. The NDWI was calculated from PlanetScope bands \cite{b9}:

\begin{equation}
\text{NDWI} = \frac{\text{Green} - \text{NIR}}{\text{Green} + \text{NIR} + \varepsilon}
\label{eq:ndwi}
\end{equation}

where $\varepsilon = 10^{-8}$ prevents division by zero. For models incorporating topographic features, HAND and slope were included alongside the four spectral bands.

\subsection{Random Forest Implementation}

RF aggregated predictions from 100 decision trees \cite{b10}. The model was trained on the single expert-annotated sample comprising approximately 50\% of pixels (stratified sampling maintaining flood prevalence). Input features for the full configuration included RGB, NIR, slope, HAND, and NDWI (7 bands total). Final predictions were obtained through majority voting across trees, with feature importance quantified using mean decrease in Gini impurity. The RF-generated flood masks served as training annotations for the subsequent U-Net models.

\subsection{U-Net Architecture}

The U-Net architecture implements a symmetric encoder-decoder structure with skip connections that preserve spatial detail while extracting multi-scale semantic information \cite{b2}. Our implementation utilized a ResNet18 backbone for the encoder, progressively downsampling through five levels while increasing channel depth from 16 to 256 filters. The decoder mirrors the encoder structure through five upsampling levels using transposed convolutions. Dice Loss was employed as the optimization criterion, particularly suited for imbalanced segmentation tasks common in flood mapping applications \cite{b11}. Training crops were resized to 128$\times$128 pixels and processed in batches. Two model configurations were evaluated: (1) optical-only using four spectral bands (RGB + NIR), and (2) full features adding topographic context (RGB + NIR + Slope + HAND = 6 bands). The training dataset consisted of 768 crops, with 320 crops reserved for testing, derived from RF-predicted flood masks covering 50\% of pixels in each handcrafted tile.

\subsection{Validation Metrics}

Performance metrics included F1-score (harmonic mean of precision and recall), IoU, precision (positive predictive value), recall (sensitivity), and overall accuracy. IoU quantifies spatial overlap particularly relevant for segmentation.

\section{Results}

\subsection{Model Performance Comparison}

Table~\ref{tab:performance} presents comprehensive performance comparison across all implemented models. The U-Net with six bands (including HAND and slope) achieved the highest performance with F1-score of 0.92 and IoU of 0.85, slightly outperforming the optical-only U-Net (F1=0.92, IoU=0.85). Both U-Net configurations substantially outperformed RF (F1=0.829, IoU=0.708), demonstrating the value of using RF-generated annotations to train DL models that can better capture spatial context through convolutional processing. The 6-band U-Net achieved higher recall (0.95) compared to the 4-band model (0.93), indicating improved flood detection capability with topographic features. However, the 4-band model showed slightly higher precision (0.91 vs. 0.89), suggesting fewer false positives. Both models achieved comparable overall accuracy ($\sim$97\%).

\begin{table}[htbp]
\caption{Performance Comparison of Different Learning Algorithms}
\begin{center}
\begin{tabular}{|l|l|l|c|c|}
\hline
\textbf{Model} & \textbf{Bands} & \textbf{Train/Test} & \textbf{IoU} & \textbf{F1} \\
\hline
Random Forest & RGB, NIR, Slope, & 50\% pixels, & 0.708 & 0.829 \\
 & HAND, NDWI & handcrafted tile & & \\
\hline
UNet (4-band) & RGB, NIR & 768/320 crops & 0.8507 & 0.9193 \\
 & & (128$\times$128) & & \\
\hline
UNet (6-band) & RGB, NIR, & 768/320 crops & 0.8520 & 0.9201 \\
 & Slope, HAND & (128$\times$128) & & \\
\hline
\end{tabular}
\label{tab:performance}
\end{center}
\end{table}

\subsection{Impact of Topographic Features}

The addition of HAND and slope features provided almost identical performance to U-Net performance, with F1-score of 0.92 and IoU of 0.85. Slight difference was in Recall, where the 6-band model achieved 0.95 compared to 0.93 for the 4-band model (+1.9\%). This suggests that HAND and slope help identify additional flooded pixels that spectral information alone might miss, likely in areas where water spectral signatures are ambiguous but topographic context indicates flooding. The trade-off is slightly lower precision (0.89 vs. 0.91), indicating some increase in false positives when topographic features are included. Feature importance analysis from RF revealed NDWI contributed 32\% of predictive power, topographic features (HAND + Slope) collectively contributed 27\%, with HAND individually accounting for 15\%. This confirms that topographic information provides complementary signals for flood classification, supporting the modest improvements observed in U-Net when these features are included.

\subsection{Visualization of Flood Inundation Extent}

Visual inspection of modeled flood inundation maps revealed that both U-Net configurations produced spatially coherent flood boundaries with minimal isolated false positives. The 6-band model showed slightly more complete flood coverage, particularly in transitional areas near water bodies where HAND values indicate flood susceptibility. RF predictions, while achieving respectable quantitative metrics, exhibited more pixelwise noise compared to the spatially continuous outputs from convolutional neural networks. The superior spatial coherence of U-Net predictions confirms the value of using RF-generated annotations to train deep learning models that can better capture flood extent patterns exhibiting spatial autocorrelation.

\begin{figure}[htbp]
\centerline{\includegraphics[width=\columnwidth]{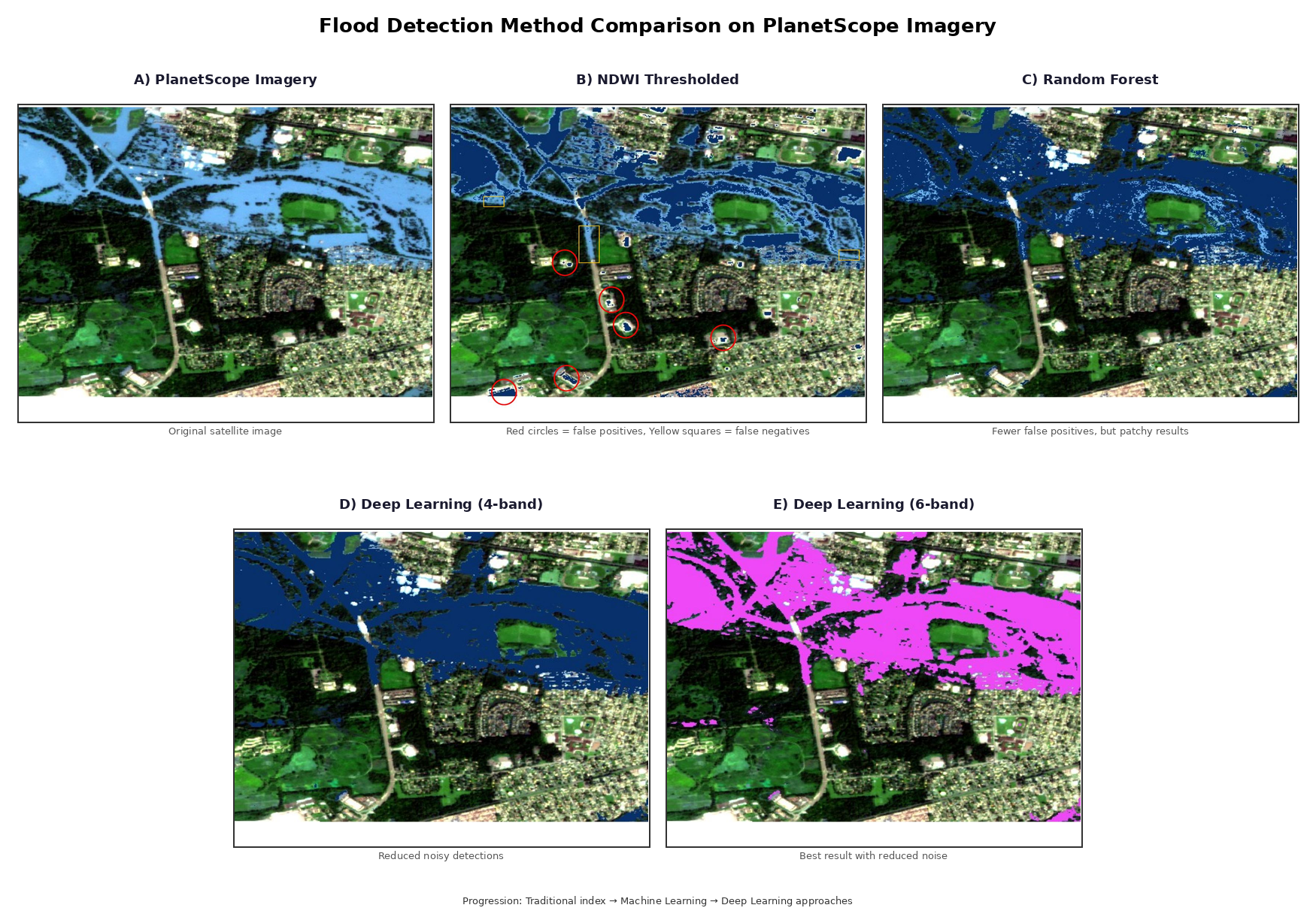}}
\caption{Flood detection method comparison on PlanetScope imagery. (A) Original satellite image, (B) NDWI thresholded results showing false positives (red circles) and false negatives (yellow squares), (C) Random Forest predictions with fewer false positives but patchy results, (D) Deep Learning 4-band with reduced noisy detections, and (E) Deep Learning 6-band achieving best results with reduced noise. The progression demonstrates improvement from traditional index-based methods to machine learning and deep learning approaches.}
\label{fig:flood_comparison}
\end{figure}

\section{Discussion and Conclusions}

This study established a progressive annotation framework where RF-generated labels are used to train U-Net models for operational flood mapping from high-resolution satellite imagery. The framework addresses a critical challenge in disaster response: the need for rapid flood maps during ongoing events when independent ground truth data are unavailable. By leveraging expert annotation scaled through Random Forest, we generated sufficient training data to train U-Net models that achieved F1-scores exceeding 0.91. The addition of with topographic features to U-Net model (6 bands) did not achieve better performance (F1=0.92, IoU=0.85 for both). This demonstrates that HAND and slope provide only marginal but positive contribution to flood detection, particularly improving recall (+1.9\%) at a small cost to precision. The Random Forest model played a crucial intermediary role, translating expert annotation into scalable training labels with 99\% accuracy on the annotated sample. This progressive strategy overcame limitations of threshold-based approaches (NDWI and OTSU) that failed to balance sensitivity and specificity in complex environments. The RF-generated annotations, while imperfect, provided sufficient training signal for U-Net to learn robust flood segmentation patterns. Feature importance analysis revealed NDWI (32\%) dominated predictions, with topographic features contributing 27\% collectively, confirming their relevance for flood classification. The value of HAND in flood mapping has been demonstrated in previous studies for rural and natural floodplains \cite{b6}, \cite{b7}, \cite{b8}. Our results extend these findings by showing that topographic features also provide modest improvements in mixed land cover settings when combined with high-resolution spectral data. The 3-meter resolution of PlanetScope imagery captures fine-scale flood patterns, and the addition of HAND helps identify inundated areas even when spectral water signatures are weak. Comparison with recent research on flood mapping supports our findings. Studies using Sentinel-1 SAR have achieved operational flood mapping with accuracies exceeding 96\% through automated thresholding \cite{b4}, while research on ML approaches has demonstrated RF achieving robust performance across diverse landscapes \cite{b12}, \cite{b13}. Our framework advances these approaches by demonstrating how ML (RF algorithm) can serve as an intermediary step to generate training annotations for DL models, enabling operational deployment in data-scarce disaster scenarios. Limitations include reliance on single-date imagery preventing temporal flood dynamics assessment, and the progressive annotation approach depending on expert interpretation quality. The PlanetScope sensor lacks shortwave infrared bands that would enable Modified NDWI for improved water discrimination \cite{b14}. Cloud cover remains a fundamental constraint for optical remote sensing of flood events. Future work should integrate Synthetic Aperture Radar, SWOT observations and hydrodynamic models to overcome optical limitations \cite{b15}, \cite{b16}, expand validation with additional flood events, and investigate alternative topographic feature representations. This study confirms that U-Net models trained on RF-derived annotations achieve robust flood inundation mapping (F1$>$0.91) from high-resolution optical imagery. The inclusion of topographic features provided marginal improvements in overall performance, but additional research is needed in other areas with distinct topography to evaluate this finding. The proposed framework offers a practical pathway for flood mapping in disaster scenarios where independent validation data are unavailable, addressing a critical need for emergency response and risk management.

\end{document}